# Patch-based Evaluation of Dense Image Matching Quality

Zhenchao Zhang [a,]*, Markus Gerke [b], George Vosselman [a], Michael Ying Yang [a]

[a] Department of Earth Observation Science, Faculty ITC, University of Twente, Enschede, The Netherlands

[b] Institute of Geodesy and Photogrammetry, Technical University of Brunswick, Germany

## Abstract

Airborne laser scanning and photogrammetry are two main techniques to obtain 3D data representing the object surface. Due to the high cost of laser scanning, we want to explore the potential of using point clouds derived by dense image matching (DIM), as effective alternatives to laser scanning data. We present a framework to evaluate point clouds from dense image matching and derived Digital Surface Models (DSM) based on automatically extracted sample patches. Dense matching error and noise level are evaluated quantitatively at both the local level and whole block level. Experiments show that the optimal vertical accuracy achieved by dense matching is as follows: the mean offset to the reference data is 0.1 Ground Sampling Distance (GSD); the maximum offset goes up to 1.0 GSD. When additional oblique images are used in dense matching, the mean deviation, the variation of mean deviation and the level of random noise all get improved. We also detect a bias between the point cloud and DSM from a single photogrammetric workflow. This framework also allows to reveal inhomogeneity in the distribution of the dense matching errors due to over-fitted BBA network. Meanwhile, suggestions are given on the photogrammetric quality control.

## 1. Introduction

Airborne laser scanning (ALS) and photogrammetry are two main techniques to obtain 3D data representing the surface of the terrain (Höhle and Höhle, 2009). Compared with airborne laser scanning, image acquisition in photogrammetry is mostly cheaper and more efficient in data acquisition flights (Hobi and Ginzler, 2012; Nurminen et al., 2013; Maltezos et al. 2016). In many countries photogrammetric image blocks are captured anyway for administrative and planning purposes with decreasing time intervals, so the question is to what extent these data can be used to replace ALS data in various application domains such as Digital Terrain Model (DTM) acquisition (Ressl et al., 2016), forestry mapping (Mura et al., 2015), classification and object extraction (Tomljenovic et al., 2016; Dong et al., 2017), and 3D modeling (Xiong et al., 2015).

In this paper, we want to explore the potential of using photogrammetric products as effective alternatives to laser scanning data (Baltsavias, 1999; Leberl et al., 2010; Remondino et al., 2014; Cavegn et al., 2014; Haala et al., 2010; Yang and Chen, 2015; Tian et al., 2017). A framework for evaluating point clouds and Digital Surface Models (DSMs) from dense image matching (DIM) using patches is proposed. The point clouds and DSMs generated from a state-of-the-art dense matching algorithm are evaluated. The quality of DIM data is evaluated at both the block level and the patch level. The contributions are as follows:

- We propose a framework for evaluating DIM point clouds and DSMs using planar patches that are automatically selected from ground point subsets. After excluding patches with possible changes between the DIM data and reference data and also patches with possible breaklines, the evaluation based on these planar patches reveals the distribution of DIM errors in the whole block for the first time. Compared with the previous point-to-point comparison, this framework is robust to local blunders and artefacts. Quantitative quality measures are designed to represent the accuracy and precision at both the local level and the whole block level.

- Concerning the oblique images, multi-view aerial photogrammetry is not yet a standard and the performance of only nadir images for dense matching is still more relevant in practice. So we also evaluate how the additional use of oblique images influences the DIM quality. We compare the quality of point clouds and DSMs from a single photogrammetric pipeline to check whether the accuracy stays the same. Meanwhile, suggestions are given on the photogrammetric quality control and dense matching parameter settings.

The paper is organized as follows: Section 2 reviews work related to 3D data evaluation. Section 3 describes the patch-based DIM evaluation framework. Section 4 presents the experiments and analyses. Section 5 discusses identified problems within the DIM data and the photogrammetric quality control. Section 6 concludes the paper.

## 2. Related work

Assessing the absolute accuracy of 3D data can be time-consuming and labor-intensive for two reasons. Firstly, the reference data must be verified as being more accurate than the compared data. Secondly, the sample size should be sufficiently large. Previous work of evaluating the absolute accuracy of 3D data can be divided into two categories based on the reference data.

Firstly, in the traditional evaluation studies, the reference data can be collected by Real Time Kinematic (RTK) GPS. However, the sample size is relatively small when the reference data are acquired by GPS device point-by-point. Jaud et al. (2016) evaluated point clouds generated from images obtained by Unmanned Aerial Vehicles (UAVs). Twenty-four ground targets were set in the study area which served as the ground control for dense matching and check points for DIM evaluation. The coordinates of these targets were obtained by post-processed differential GPS. Hobi and Ginzler (2012) evaluated the quality of DSMs from stereo matching of WorldView-2 satellite images and ADS80 aerial images using 36 reference points obtained by sub-decimeter GPS after differential correlation. Nurminen et al. (2013) studied the accuracy of DSMs derived from ALS and DIM in the estimation of plot-level variables. The reference variables of the forest plots were obtained by field surveys. Gerke et al. (2016) studied the impact of GCP distribution, use of oblique images or not, different software on the overall geometry of the block. Their conclusions were drawn according to the Root Mean Square Errors (RMSEs) of GCPs and CPs in bundle block adjustment (BBA). However, they didn't further look into the quality of the final DIM points. No research has verified that low RMSEs of check points in BBA are equivalent to accurate dense matching data.

Secondly, the reference data may be obtained by laser scanning. The basic assumption is that the point clouds obtained by laser scanning are more accurate than point clouds from photogrammetry, at least concerning the height component. Mandlburger et al. (2017) calculated the deviation between DIM-DSM and Lidar-DSM at impervious surfaces and found a systematic deviation of 4.3 cm and a dispersion of 4.1 cm. Tian et al. (2017) selected 184 inventory plots as the samples for DSM evaluation in a forest area. Two datasets from ALS were taken as reference data. Similar work taking laser scanning data as reference can also be found in (Poon et al. 2005; Kraus et al., 2006; Gehrke et al., 2008; Moussa et al., 2013; Remondino et al. 2014; Nex et al., 2015; Jaud et al., 2016; Maltezos et al., 2016; Sofia et al., 2016; Ressl et al 2016).

Some studies evaluated the point cloud derived from Semi-Global Matching (SGM) by making comparisons with ALS data or TLS data on a planar sports field, complex castle or building façade (Rothermel et al, 2012; Haala and Rothermel, 2012; Cavegn et al., 2014; Remondino et al., 2017). However, the small sample size or local area cannot represent the error distribution in the whole block. When calculating quality measures, point-to-point distance (Nex et al., 2015) and point-to-plane distance (Rothermel and Haala, 2011) are widely used as the measures to represent the bias. However, these measures are sensitive to blunders and random noise in the point cloud. In addition, they are less reliable or persuasive if calculated without consideration of the breaklines in natural scene, e.g. the bumpy terrain, the edges of traffic islands or curbstones, and edges and ridges of roofs (Haala, 2015; Ressl et al., 2016; Jaud et al., 2016).

In our previous work of evaluating point cloud from multi-view photogrammetry (Zhang et al., 2017), robust quality measures were calculated on each roof segment. The problem was that the roof sizes and inclination angles varied from roof to roof. Therefore, in this paper, the DIM quality is evaluated based on planar patches of the same size extracted from smooth ground. This allows the generation of a large sample for robust quality evaluation and statistical analysis.

## 3. Methodology

In our evaluation framework, the point cloud from airborne laser scanning is taken as reference data. The ALS data are assumed to be accurate with regards to the external reference and precise in consideration of random noise. The "patches" are regular squares selected on the smooth ground from the ALS data. Every patch is a sample for quality evaluation. Therefore, the large sample size can indicate the error distribution in the whole block. The proposed framework for DIM evaluation includes two steps: first, detect square patches from the ALS data (See Fig. 1); second, select the neighboring DIM points for each patch and calculate the quality measures.

### 3.1. Patch detection

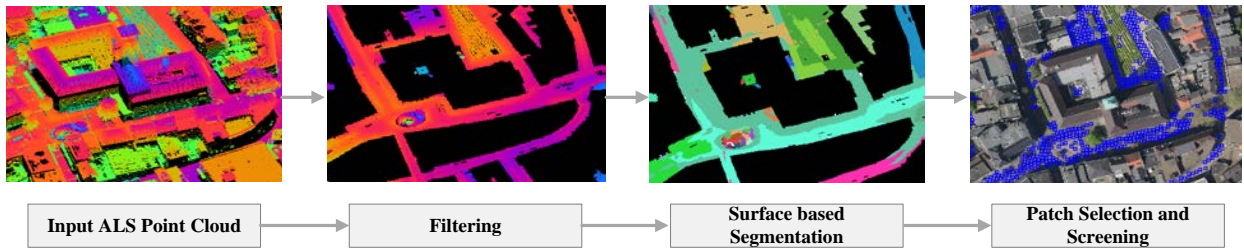

Fig. 1. Workflow for detecting square patches from ALS point cloud.

The goal of patch detection is to localize valid patches on the ALS data. The patches taken as samples should be selected from the smooth ground area in the ALS data. The selection of patches should avoid data gaps, breaklines and strong noise. Planar patches of uniform size with acceptable noise level will be considered valid. As shown in Fig. 1, the workflow first identifies ground points from the ALS data using the method of (Axelsson, 2000). Then planar segments are extracted from ground points using a surface-based growing method (Vosselman, 2013). This approach employs the 3D Hough transform to detect seed surfaces. Then the nearby points are added to the surface if the distance from a certain point to the fitted plane is below a certain threshold. After new points are added to the segment, the plane parameters are recalculated before testing the next point. Slight over-segmentation is preferred over under-segmentation: over-segmentation will ensure better planarity and help avoiding breaklines in the segments.

After segmentation, the laser points with segment labels should be screened to discard small clusters or noisy points before patch selection. Features listed in Table 1 are used to remove these small or noisy segments. *Segment size* is used to eliminate small segments; *linearity of segment* is used to eliminate narrow segments; *Normal slope* is used to exclude segments on a steep slope; *average angle* and *residual of plane fitting* (*RPF*) are used to eliminate noisy clusters. A segment is kept only if it passes the check based on the five features.

Table 1. Segment-based features for roof segment extraction

| Feature | Description |
| --- | --- |
| *Segment size* | Number of points in the segment |
| *linearity of segment* | $(\lambda_1 - \lambda_2)/\lambda_1$ , $\lambda_1$ is the maximum eigenvalue of the covariance matrix (Weinmann et al., 2015) |
| *Normal slope* | Normal direction of the fitted plane |
| *Average angle* | Mean of the angles between local point normals and fitted plane normal |
| *Residual of plane fitting* (*RPF*) | Standard deviation of the distances between points and the fitted plane in a segment |

Patch selection is implemented in the bounding box of the horizontal segments. Fig. 2 shows that the bounding box is calculated around all the points in the segment. A grid is built within the bounding box in the horizontal space. If there is no point within a certain grid cell, the grid cell is set to *empty* (i.e. white cells in Fig. 2).

When selecting patches, the algorithm starts from the corner of the *grids* and iterates over each cell. A patch is compiled out of several initial grid cells and is within the bounding box of the segment. If no data gap is detected in any cell within this patch, this patch will be valid. In this way, the patch selection method can automatically avoid the locations of data gaps in the segments. Additionally, in order to speed up the iteration, the stride can be two or more grid cells each time. Due to the "brute-force" search over dense grid cells, many selected patched are overlapping. These overlapping patches need screening. Square patches of uniform size are preferred because its regular shape makes the patch selection and attribute calculation much easier.

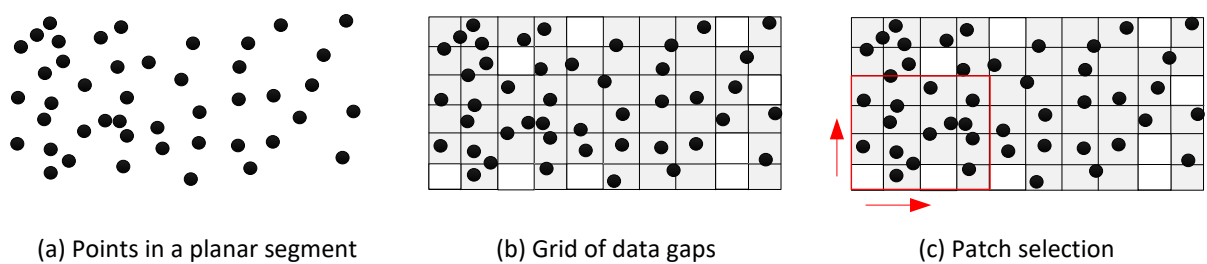

(a) Points in a planar segment  (b) Grid of data gaps  (c) Patch selection

Fig. 2. Patch selection from the data gap grid in the horizontal space. In (b) and (c), the white cells indicate data gaps or empty cells, the grey cells indicate non-data gap. (c) shows that the patch size is 4 × 4 cells: the patch will be valid only if there is no data gap in the 16 cells.

### 3.2. Patch-based quality measures and screening

After patch selection, the valid locations for patches are obtained. The DIM points of a certain patch are selected from the neighbouring points of the selected ALS patch. The selection of DIM points adopts the same bounds as the ALS patch. Then the quality measures are calculated for every patch.

When evaluating the quality of 3D data, almost all the previous DIM evaluation papers studied the accuracy of 3D data. Rothermel and Haala (2011) and Leberl et al. (2010) also looked into the point cloud density. This paper evaluates two factors related to the data quality:

- Accuracy: the deviation between the compared data and the ground truth (or reference data).

- Precision: the relative closeness of many measurements to each other, a.k.a. the level of random noise.

Accuracy and precision are independent of each other. This paper only focuses on the vertical accuracy of the point cloud and DSM. Assuming that the 3D data show a normal distribution and contain no blunders, Table 2 shows the quality measures calculated at the patch level and the photogrammetric block level to represent the data accuracy and precision. The patch-based measures are aggregated into the block-level quality measures. In Table 2, $i$ denotes the index of a patch in the whole block; $j$ denotes the index of a specific DIM point in a certain patch. $\Delta h_{ij}$ denotes the deviation from the $j$ th DIM point to the plane which is fitted from all the ALS points within the $i$ th patch. $n_i$ denotes the number of DIM points in the $i$ th patch. $m$ denotes the number of patches in the whole block which is also the sample size for statistical analysis.

Table 2. Quantitative quality measures for dense matching evaluation

| Level | Quality Measure | Definition | Meaning |
|---|---|---|---|
| Patch-level | Mean deviation | $\mu_i = \frac{1}{n_i}\sum_{j=1}^{n_i} \Delta h_{ij}$ | Mean deviation of DIM points in a certain patch w.r.t. the reference ALS plane |
| | Standard deviation | $\sigma_i = \sqrt{\frac{1}{n_i - 1}\sum_{j=1}^{n}(\Delta h_{ij} - \mu_i)^2}$ | Precision of DIM data in a certain patch |
| Block-level | Mean of mean deviations (M_MD) | $\bar{\mu} = \frac{1}{m}\sum_{i=1}^{m} \mu_i$ | Average deviation of DIM data from the reference data in the whole block |
| | Standard deviation of mean deviations (STD_MD) | $\sigma_{\mu_i} = \sqrt{\frac{1}{m-1}\sum_{i=1}^{m}(\mu_i - \bar{\mu})^2}$ | Noise level of the mean deviations at the block level |
| | Average standard deviation (A_STD) | $\mu_{\sigma_i} = \sqrt{\frac{1}{m}\sum_{i=1}^{m} \sigma_i^2}$ | Average precision of the DIM data in the block |

In Table 2, *Mean deviation* and *Standard deviation* are the quality measures at the patch-level. Both *M_MD* and *STD_MD* are measures indicating the accuracy at the block level. Specifically, a larger *STD_MD* indicates more dispersed patch-based errors in the block. *Standard deviation* is the measure for representing the data noise level. *A_STD* is to indicate the level of random noise in the whole block.

### 3.3. Patch screening

Rule-based screening are implemented again at the patch level (previously implemented at the segment level in Section 3.1). Five rules are employed to screen the patches.

(1) *Number of points* in the DIM patch: check based on Number of points is used to make sure there are sufficient points in the DIM patches. So those patches with severe data gaps will be eliminated.

(2) *Mean deviation*: to make sure that the mean deviation is caused by dense matching error but not changes between the ALS data and DIM data. The Mean deviation should be smaller than the maximum mean deviation.

(3) *Shading attribute*: Empirically, the DIM quality in shaded and non-shaded area are different so patches under shadow should not be used for evaluation. The shadow mask is calculated from an orthoimage based on a grayscale histogram (Sirmacek and Unsalan, 2009). Only if all the four corners and the center location of a certain patch lie in the non-shaded area, the patch is taken as non-shaded patch and thus be used.

(4) *Green index*: After the above screening, some patches on the grassland can still be left. The *Normalized Excessive Green Index* (nEGI) in Eq. (1) is used to filter out vegetation patches on the orthoimages (Qin, 2014).

$$nEGI = (2G - R - B)/(2G + R + B) \qquad (1)$$

Similar with *Shading attribute*, only if all the four corners and the center of a patch are labeled as non-vegetation, this patch will be used for evaluation.

(5) *Overlapping attribute*: In order to keep as many valid patches as possible, filters out redundant patches is implemented at the last step of patch screening. Some patches are overlapping in space due to the previous dense selection from the data gap grid (see Fig. 2). The densely overlapping patches will be screened automatically based on the spatial relationship to make sure a certain location in the study area will only be used once.

After patch screening, the remaining patches with sufficient DIM points are densely distributed on the open ground. DIM evaluation and statistical analyses are performed based on these valid patches.

## 4. Materials and Results

### 4.1. Study area and experimental setup

The study area is located in Enschede, The Netherlands. The coverage area by aerial images is larger than the quality evaluation area. Fig. 3 shows the area for quality evaluation (1.6 km$^2$). This area is a densely-built urban area mainly covered by buildings, roads, squares, railways and vegetation. The ALS data were acquired in 2007 and the maximum systematic error in height is 5 cm (Vosselman, 2008). 510 aerial images including 102 nadir images and 408 oblique images were obtained by *Slagboom en Peeters* in 2011 together with exterior orientations. The tilt angle of oblique view is approximately 45°. The image size is 5616 × 3744 pixels. The GSD of nadir images equals 10 cm. The overlap of nadir images is approximately 75% both along track and across track. In the block, ground reference targets (RTs) were measured with a Leica CS15 receiver using real time kinematic GPS. The 3D accuracy of all RTs were set to be better than 3 cm during GPS measurement. All of the reference targets are the corners of the zebra crossings, centers of manholes or other distinctive corners in the urban scene.

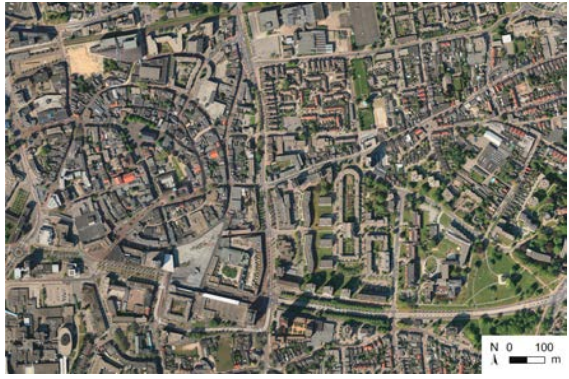

Fig. 3. Orthoimage of the area for quality evaluation. This orthoimage is generated by Pix4D.

Before taking ALS data as reference, we should check the level of possible systematic errors in the ALS data. Taking the RTs as ground truth, the ALS data can be compared to the RTs. Since all the RTs are located in the open area so planes can be fitted using neighboring ALS points. The vertical residual from a RT to the fitted ALS plane is calculated as the indicator for the ALS accuracy. Results show that the mean deviation ($\mu$) and standard deviation ($\sigma$) between the RTs and the fitted ALS plane are 0.013 m and 0.031 m. Furthermore, if the residual from RT to the ALS fitted plane is larger than three times of the standard deviations ($\sigma$), this RT will be discarded. This cross-verification ensures that both the ALS data and RTs used in the BBA, dense matching and DIM evaluation are reliable. Finally, 99 RTs passing this cross-verification are used as GCPs and check points in the BBA.

### 4.1.1. Bundle adjustment (BBA)

In the step of BBA, two configurations with 5 and 44 GCPs are set up for comparison. "5 GCPs" indicate the configuration with only a few ground references; while 44 GCPs indicate the case with many ground references. The distribution of GCPs are shown in Fig. 4. Since the GCP distribution will also influence the BBA accuracy (Gerke et al., 2016), the GCPs are evenly selected from the block. When 5 or 44 RTs are used as GCPs, the remaining 94 and 55 RTs are taken as check points, respectively. The results of the two configurations with 5 GCPs and 44 GCPs are presented in Section 4.2 and Section 4.3, respectively.

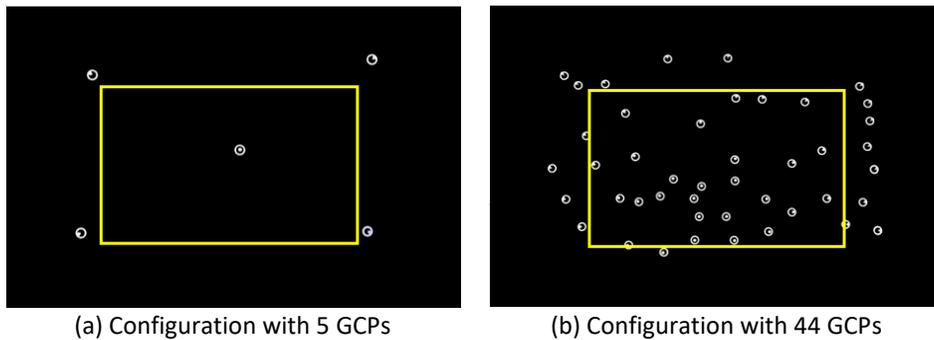

(a) Configuration with 5 GCPs      (b) Configuration with 44 GCPs

Fi.g.4. Two configurations with different GCP amounts used in bundle adjustment. The white dots show the distribution of GCPs in the block. The yellow rectangle indicates the area for DIM evaluation (1.6 km$^2$).

The BBA is run in Pix4Dmapper Pro (version 3.2). Image scale for BBA is kept constant at the "original scale" so the original resolution of images is used in BBA. The horizontal accuracy of the 3D GCPs is set to "2 cm (default value)" which controls the GCPs weights inn BBA. Table 3 shows the vertical RMSEs of GCPs and CPs.

Table 3. Vertical RMSEs of GCPs and CPs when horizontal accuracy of GCPs is set to 2 cm

| Number of GCPs | Number of CPs | RMSE of GCPs (m) | RMSE of CPs (m) |
| --- | --- | --- | --- |
| 5 | 94 | 0.007 | 0.060 |
| 44 | 55 | 0.018 | 0.031 |

### 4.1.2. Dense image matching

For the execution of dense image matching, we select the state-of-the-art software SURE (Surface Reconstruction, version 2.1.0.33) from nFrames. A few researches have reported its performance in data accuracy (Haala and Rothermel, 2012; Rothermel et al., 2012; Ressl et al., 2016). The dense matching algorithm in SURE is a tube-shaped SGM (t-SGM). The SGM method in (Hirschmüller, 2008) is improved by restricting the disparity searching space which leads to a higher efficiency. Furthermore, the redundant disparity information is exploited to eliminate blunders and increase the accuracy of depth.

SURE is software for dense matching (not for BBA) whose interior orientation (IOs) and exterior orientation (EOs) elements can be imported from Pix4D. Several parameters are supposed to affect the dense matching quality. *Minimum Model Count* (*MMC*) represents the minimum number of models for a 3D point to be considered valid during triangulation. A larger *MMC* will increase the reliability of generated points but also lead to a lower number of accepted matched points. *MMC* is fixed to 2 in all our experiments. When *MMC* is set larger (e.g. 3), there will be many data gaps in the narrow alley. The image scale for dense matching is fixed to 1/2. The interpolation method for DSM generation is set to *Inverse Distance Weighting* (IDW).

Then we evaluate the DIM quality based on the configuration with 5 GCPs to answer the following two research questions: (1) What is the impact of additional use of oblique images on the DIM accuracy and precision? (2) Is the accuracy of point cloud and DSM from a single photogrammetric pipeline the same? In addition to the point cloud from dense matching, a DSM is also obtained from a standard photogrammetric workflow. The DSMs generated by SURE can be grid data structure but saved in point cloud format. Same with cropping point cloud patches, the DSM patches are generated by cutting out the corresponding patch area from the raster DSM. The resolution of the DSM grid is 10 cm. Four data sets are obtained:

(1) GCP05_N+O_PC;    (2) GCP05_N+O_DSM;

(3) GCP05_N_PC;    (4) GCP05_N_DSM.

The naming scheme of the four data above shows different parameters of the data. GCP05 means 5 GCPs are used in BBA; N+O indicates that both nadir (N) and oblique (O) images are used in dense matching; "N" indicates that only nadir images are used in dense matching; PC or DSM means this is point cloud or DSM. Note that DSM (2) is interpolated from point cloud (1); (4) is generated from (3).

### 4.1.3. Parameter settings for DIM evaluation

In the patch selection, a surface growing radius of 1.0 m and maximum distance between point and fitted plane of 0.2 m are employed. In addition, the size of each cell in Fig. 2(c) is set to 0.5 m for both data sets. The patch size is set to 2 m × 2 m (i.e. 4 × 4 cells ). The thresholds for the rules in Table 1 are set as below: *segment size* is 100, *linearity of segment* is 0.99, *normal slope* is 45°, *RPF* is 0.1 m, *average angle* is 5°. In patch screening, the threshold for *Number of points* is determined from the histograms of the point amounts in a DIM patch in the whole block to keep sufficient points. The Mean deviation threshold is set by adding the 99% percentile of the mean deviation in the block with some small tolerance value (e.g. 2 cm). The threshold for nEGI in Eq. (1) is set to 0.1.

## 4.2. Results of the configuration with 5 GCPs

After patch selection and screening, the ground patches are classified into non-shaded bare ground, shaded ground and grassland. Some examples of the extracted patches are shown in Fig. 5. Fig. 5 shows that the non-shaded bare ground, grassland and shaded ground are correctly distinguished. The right figure shows the selected patches on the central bus station of Enschede. The white stripes are actually platforms higher than the grey ground by around 20 cm. The proposed algorithm performs well in extracting planar patches as well as avoiding breaklines.

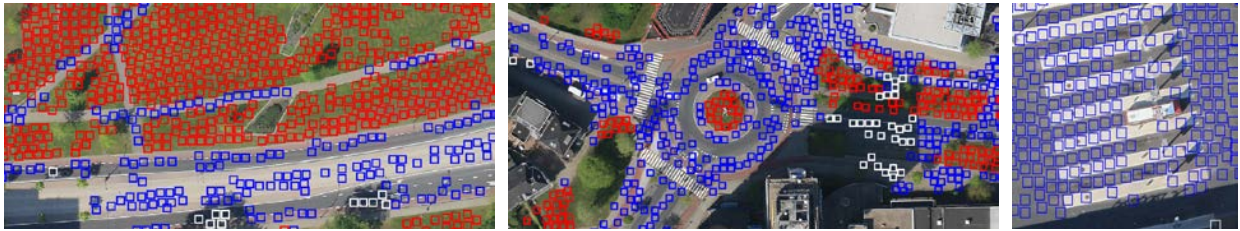

Fig. 4. Examples for extracted patches. Patch size is 2 m × 2 m. Different land cover types are distinguished by colors: blue indicates non-shaded bare ground; red indicates grassland, white indicates shaded ground. (For interpretation of the references to color in this figure, the reader is referred to the web version of this article.)

Finally, 24,634 non-shaded patches of 2 m × 2 m are selected on the ground, i.e. only the blue patches in Fig. 4. In order to make the block-level quality measures comparable, the amounts and locations of the patches used for evaluating the four data sets are all the same.

Table 4 shows the quality measures at the block level calculated for the four data sets. Comparing the first row shows the impact of oblique images on the DIM quality; comparing the first row with the second row shows the difference between point cloud and DSM in accuracy and precision level. The two questions are discussed separately below.

Table 4. Quality measures for point clouds and DSMs in configurations with 5 GCPs (Unit: m). The three quality measures in each cell are $M\_MD$, $STD\_MD$, $A\_STD$ from left to right.

|  | N+O | N |
| --- | --- | --- |
| Point cloud (PC) | 0.002; 0.040; 0.094 | 0.016; 0.045; 0.106 |
| DSM | 0.034; 0.060; 0.048 | 0.024; 0.066; 0.083 |

### 4.2.1. Evaluation of the impact of oblique images

The first row of Table 4 shows the comparison between GCP05_N+O_PC and GCP05_N_PC. A general finding is that when both nadir and oblique images are used for dense matching, all the three quality measures are better than the measures of configuration with only nadir images. The $M\_MD$ gets improved remarkably by 1.4 cm from 0.016 m to 0.002 m when oblique images are used; The $STD\_MD$ is improved very slightly by 0.5 cm; The $A\_STD$ is improved by 1.2 cm.

The distribution of mean deviations in the whole block for the two configurations are shown in Fig. 5. A normal distribution is estimated using the mean and standard deviation calculated from the same data set. The normal distribution is stretched and then superimposed on the histogram to visualize the deviation between the real measurements and a normal distribution (Höhle and Höhle, 2009). In each histogram, the horizontal axis indicates the patch-based mean deviation, the vertical axis indicates the frequency in the whole block. The scale and interval of the axes for the two histograms are all the same.

The peak of Fig. 5(a) is located at approximately 0 which corresponds with $M\_MD$=0.002 m in Table3. The histogram is centralized and "thin" in shape which corresponds with $STD\_MD$=0.040 m. The mean deviations range from -0.09 m to 0.10 m which means that in almost every patch, the vertical error of dense matching is better than 1 GSD. Fig. 5(b) shows a relatively dispersed histogram compared to Fig. 5(a). And in Fig. 5(a), the peak is locate at 0.016 m at the horizontal axis. The range of mean deviations from -0.10 m to 0.12 m is a little worse than Fig. 5(a).

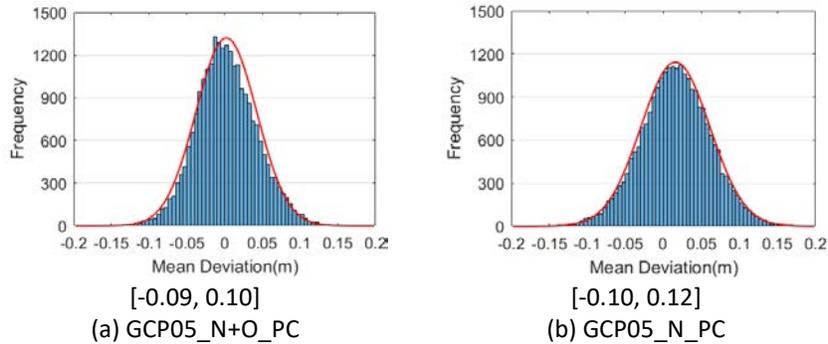

[-0.09, 0.10]                    [-0.10, 0.12]
(a) GCP05_N+O_PC            (b) GCP05_N_PC

Fig. 5. Distribution of mean deviations for 24,634 non-shaded ground patches. For the naming scheme of the configurations, please refer to Section 4.1.2. All the histograms are overlaid with an estimated stretched normal distribution. The interval below each histogram is taken from 0.01 quantile to 0.99 quantile of the histogram.

Fig. 6 shows the distribution of patch-based standard deviations from the whole block. Both the two histograms are close to a normal distribution. The patch-based standard deviations range from approximately 6 cm to 15 cm in the block. The two histograms show slight difference in shape which corresponds to 1.2 cm difference in Table 4.

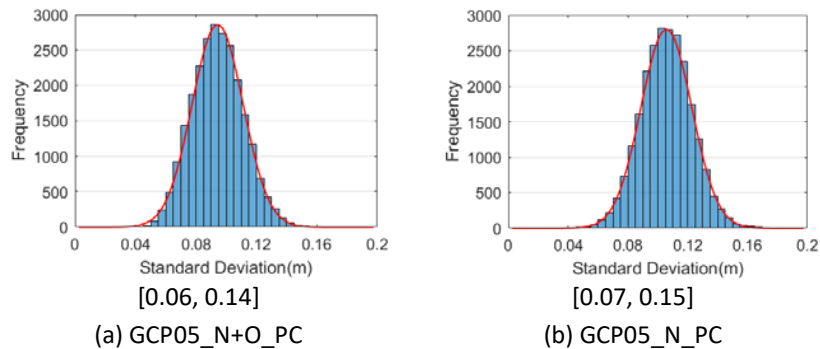

[0.06, 0.14]                    [0.07, 0.15]
(a) GCP05_N+O_PC            (b) GCP05_N_PC

Fig. 6. Distribution of standard deviations for 24,634 non-shaded ground patches. The interval below each histogram is taken from 0.01 quantile to 0.99 quantile of the histogram.

Fig. 7 shows the patch-based mean deviations in the whole block for the data GCP05_N+O_PC colored according to the absolute mean deviation values. Generally, the DIM errors are homogenous in the whole block. In some locations, especially along narrow alley, the mean deviations may get worse.

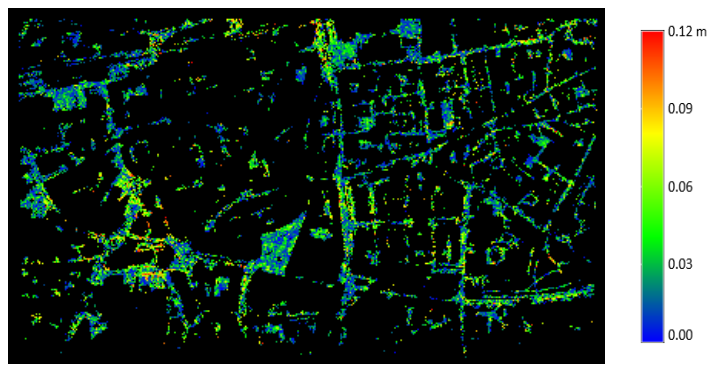

Fig. 7. Patch-based mean deviations in the whole block colored by the absolute values for the data GCP05_N+O_PC. Color coding from blue to red indicates that the mean deviation gets larger and larger. (For interpretation of the references to color in this figure legend, the reader is referred to the web version of this article.)

### 4.2.2. Comparison between point cloud and DSM

As expected, the *A_STD* of DSMs indicating the noise level is much lower than the point clouds since interpolation is employed from point clouds to DSMs. Based on our evaluation framework, we observe a bias between point cloud and DSM from the SURE pipeline. The first column in Table 4 shows that the difference of *M_MD* between point cloud and DSM for N+O is 3.2 cm, which means generally the DSM surface is higher than the point cloud surface by 3.2 cm. The second column in Table 4 shows that the DSM surface is generally higher than the point cloud surface by 0.8 cm when only nadir images are used in dense matching. The interpolation process changes the accuracy of the data and the impact level depends on the point cloud density. A recent paper (Mandlburger et al, 2017) also reported the same deviation between point cloud surface and DSM surface so this deviation might be caused by the interpolation process in the software. In order to further check the deviation between point clouds and DSMs, Fig. 8 shows the overlaid histograms of mean deviations for point clouds and DSMs.

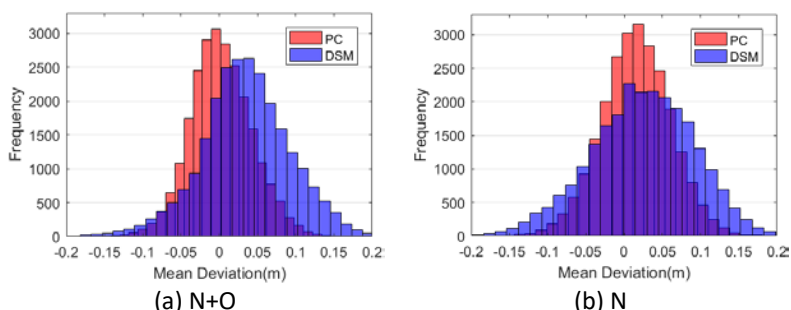

(a) N+O  (b) N

Fig. 8. Overlaid histograms of mean deviations for point cloud and DSM. Red histograms indicate the distributions of point clouds; blue histograms indicate the distributions of DSMs. "PC" in the legend indicates point cloud. (For interpretation of the references to color in this figure, the reader is referred to the web version of this article.)

Fig. 8(a) shows a clear deviation between the peaks of the two histograms while in Fig. 8(b) the two peaks are relatively close to each other. Another observation is that the distribution of mean deviations for DSMs is more dispersed than the point cloud, which corresponds to Table 4 that the *STD_MD* of DSMs is larger than point clouds. In summary, comparing point cloud and DSM, although the noise level is reduced, the absolute accuracy is changed during interpolation and the error distribution for DSMs is more dispersed than point cloud.

### 4.2.3. Visualization of patch-based mean deviations

The patch-based mean deviations are visualized in Fig. 9. This square paved by concrete in downtown Enschede is relatively smooth. The patches are filled with the value of quality measures and colored based on positive or negative values.

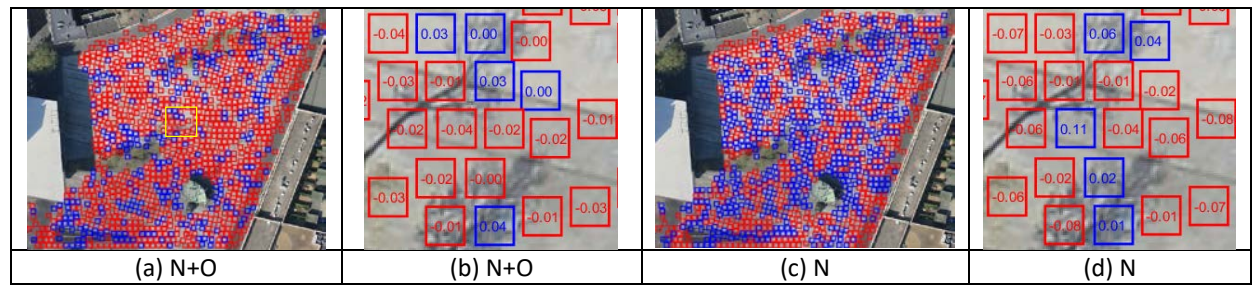

(a) N+O  (b) N+O  (c) N  (d) N

Fig. 9. Visualization of patch-based mean deviations for the data sets GCP05_PC. The blue patches indicate positive values while the red indicates negative values. (a) and (b) show the mean deviations for the data GCP05_N+O_PC, (c) and (d) show the quality measures for GCP05_N_PC. The yellow rectangle on (a) indicates the area for visualization from in (b) and (d). (For interpretation of the references to color in this figure, the reader is referred to the web version of this article.)

Fig. 9(a) and (c) show that the patch-based mean deviations vary between positive and negative on the square. Therefore, the point cloud surface from dense matching is fluctuating around the referred ALS surface. In addition, comparing Fig. 9(a) and (c), or (b) and (d) shows that whether or not oblique images are used in dense matching makes a big impact on the local patch-based accuracy.

### 4.3. Overfitting in the configuration with 44 GCPs

We evaluate the point cloud *GCP44_N+O_PC* using the same 24,634 patches as we did for the configuration with 5 GCPs. The quality measures are shown in Table 5. It is clear that both the *M_MD* and *STD_MD* get worse when more GCPs are used.

Table 5. Quality measures for point clouds when the GCP weights in BBA is set to 2 cm (Unit: m). The three quality measures in each cell are *M_MD*, *STD_MD*, *A_STD* from left to right.

| Configuration | Quality measures |
|---|---|
| GCP05_N+O_PC | 0.002; 0.040; 0.094 |
| GCP44_N+O_PC | -0.026; 0.049; 0.098 |

However, Table 3 in Section 4.1.1 shows when the GCP amounts increases from 5 to 44, the RMSE of check points in BBA reported by Pix4D decreases from 6.0 cm to 3.1 cm. Thus the check points indicate that the accuracy of BBA is getting better when more GCPs are used. Previous studies e.g. (Gerke et al., 2016) have also verified that the more GCPs, the better the BBA accuracy will be.

Therefore, when the GCP amount increases from 5 to 44, the accuracy of BBA gets improves but the accuracy of the point cloud deteriorates. These contradictory findings can be further checked by visualizing the patch-based mean deviations in the whole block as shown in Fig. 10.

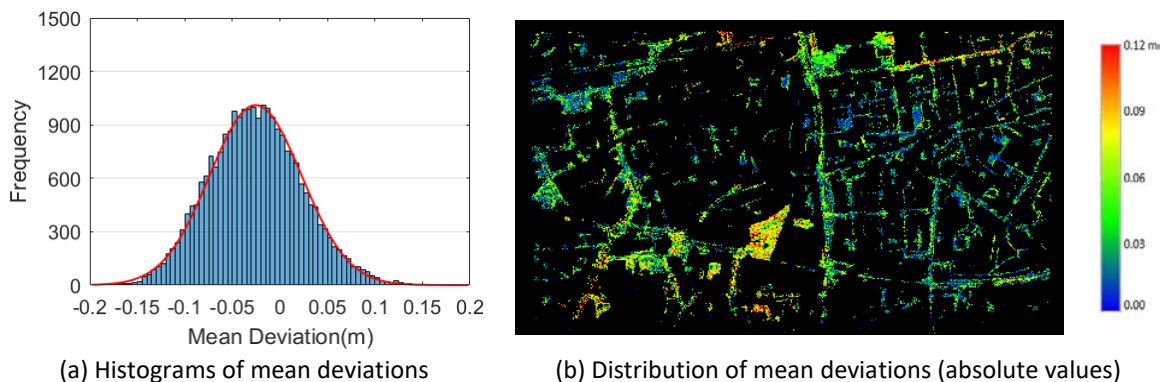

(a) Histograms of mean deviations       (b) Distribution of mean deviations (absolute values)

Fig. 10. Distribution of mean deviations for GCP44_N+O_PC when GCP weight for BBA is set to 2 cm. (For interpretation of the references to color in this figure legend, the reader is referred to the web version of this article.)

Fig. 10(a) shows that the peak of this histogram is negative (i.e. M_MD=-0.026 m). The mean deviations range in [-0.13 m, 0.09 m] which is much worse than the configuration with 5 GCPs. Fig. 10(b) shows inhomogeneous error distribution in the block. The absolute mean deviations in the southwest of the block (i.e. the third quadrant) is generally larger than other parts. The block is thus verified to be "overfitting". The reason might be that many GCPs (44) with high weights make the BBA network stiff. When the BBA network is "overfitting", the dense matching error will also show obvious inhomogeneity in the block. Furthermore, in order to check whether the BBA network is "overfitting", the GCP weights are set to 5 cm in the BBA in Pix4D, the RMSE of GCPs is 0.019 m, the RMSE of CPs is 0.031 m. Compared with Table 3, the RMSEs of GCPs and CPs change very slightly.

Then we evaluate the new point cloud generated by SURE with new orientations. We find a large improvement in the point cloud quality. The *M_MD*, *STD_MD*, *A_STD* for the new point cloud from GCP weights of 5 cm is 0.011 m, 0.044 m and 0.094 m. And the "overfitting" effect is alleviated as shown in Fig. 11.

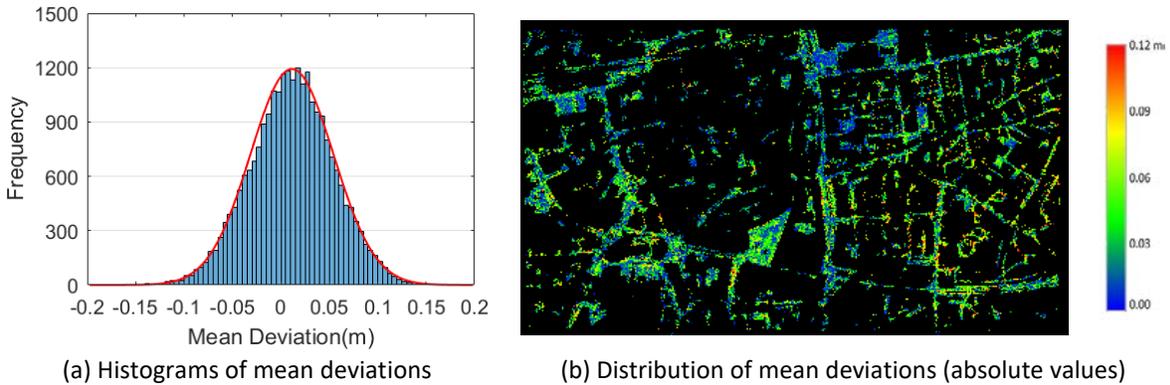

(a) Histograms of mean deviations  (b) Distribution of mean deviations (absolute values)

Fig. 11. Distribution of mean deviations for GCP44_N+O_PC when GCP weight for BBA is set to 5 cm. (For interpretation of the references to color in this figure legend, the reader is referred to the web version of this article.)

The mean deviations range in [-0.09 m, 0.11 m] in Fig. 11(a). Fig. 11(b) shows the homogeneity level of mean deviations in the block is much better then Fig. 10(b) and no obvious "over-fitting" appears.

## 5. Discussion

In our evaluation framework, both the M_MD and STD_MD to represent the dense matching accuracy in the block. The M_MD indicates the general bias of the DIM points from the reference; STD_MD indicates the dispersion level of the dense matching errors. In Section 4.2.1, when oblique images and nadir images are both used in dense matching, the M_MD gets improved, but the STD_MD keeps relatively stable.

A good point cloud should not only be accurate but also represent the object details with little random noise. When tuning the parameters in dense matching, the key is to balance between data gap level and noise level. Dense matching quality depends largely on the image contrast and texture. In order to obtain less noisy points from the SURE software on the narrow streets or under shadow, the parameter MMC should be set as large as possible as long as the data gap level is still acceptable.

Finally, even if the ALS accuracy is verified in Section 4.1, we should still consider the uncertainty of ALS data. The results in Section 4 will be influenced if there is a systematic error of 0.013 cm within the ALS data. Assuming that the ALS errors are homogeneous, the values for M_MD will be changed. The histograms in Fig. 5 will be translated towards the right side by 0.013 m. In this case, the overfitting of the BBA network with 2 cm GCP weights still occurs. In addition, the three measures: patch-based standard deviation, STD_MD, A_STD will not be influenced by the uncertainty of the ALS data for two reasons: firstly, a plane is fitted to local ALS plane as reference for DIM evaluation so the influence of local noise in ALS data is small; secondly, the mean deviations or M_MD are subtracted from these measures so the accuracy of ALS data will not influence these measures.

Concerning the "overfitting" in the BBA, this effect is not detected by the check points but can be viewed in the point cloud based on our evaluation framework. Many previous studies focused on evaluating the BBA accuracy based on the RMSEs of check points. Our finding shows that the RMSEs of check points in the BBA are not equivalent to the point cloud accuracy from dense matching. In addition, the BBA network will get "stiff and overfitting" when many GCPs given high weights are employed in BBA. On the other hand, the point cloud GCP05_N+O_PC with only 5 GCPs has already achieved high point cloud accuracy. Actually, when only a few GCPs are used, the BBA will become more sensitive to the selection of GCPs. In this case, the BBA network is easier to get biased due to one or two inaccurate GCPs.

# 6. Conclusion

We have presented a framework for evaluating the quality of 3D point clouds and DSMs delivered by dense image matching in urban area. Planar patches of uniform size are extracted from smooth terrain with the guidance of ALS data. The traditional evaluation work based on check points simply reveals the BBA accuracy. In contrast, our evaluation framework based on large sample size reveals the distribution of dense matching errors in the whole photogrammetric block for the first time. This framework is also robust to possible blunders and artefacts in the DIM points. Robust quality measures are proposed to represent the DIM accuracy and precision quantitatively. Experiments show that the optimal accuracy of DIM point cloud is as follows: the M_MD is better than 0.1 GSD; the STD_MD is better than 0.5 GSD; the maximum mean deviation reaches 1.5 GSD. It means that on almost all the patches, the dense matching errors are better than 1 GSD.

We also verify at the whole block level that when oblique images are used in dense matching together with nadir images, the accuracy will get improved and the noise level will decrease. The evaluation framework also reports a deviation between the point cloud and DSM from a single photogrammetric workflow. The deviation will be less distinct when the point cloud density drops. When many GCPs with high weights are employed in BBA, the BBA network may get "overfitting" which is reflected in the inhomogeneous distribution of the patch-based DIM errors. Interestingly, this problem cannot be discovered by using check points. Since dense matching is time-consuming and involves many factors and parameters, future work can still evaluate the impact of other factors (e.g. GCP amounts, image scale, MMC) on the point cloud quality.